\UseRawInputEncoding 
\documentclass[review]{elsarticle}

\usepackage{url,hyperref,microtype,subcaption}
\usepackage[onehalfspacing]{setspace}
\usepackage{graphicx}
\usepackage{epstopdf}
\usepackage[switch]{lineno}
\usepackage{amsmath}
\usepackage{algorithm}
\usepackage{algorithmicx}
\usepackage{algpseudocode}
\usepackage{caption}
\modulolinenumbers[5]

\journal{Journal of \LaTeX\ Templates}

\begin{document}

\begin{frontmatter}

\title{Brain-inspired Graph Spiking Neural Networks for Commonsense Knowledge Representation and Reasoning}

\author{Hongjian Fang\fnref{a,c,e}}
\ead{fanghongjian2017@ia.ac.cn}

\author{Yi Zeng\fnref{a,b,c,d,e,f}}
\ead{yi.zeng@ia.ac.cn}

\author{Jianbo Tang \fnref{a}}
\author{Yuwei Wang \fnref{a}}
\author{Yao Liang  \fnref{a}}
\author{Xin Liu \fnref{a}}

\fntext[a]{Research Center for Brain-inspired Intelligence, Institute of Automation, Chinese Academy of Sciences, Beijing, China}
\fntext[b]{Center for Excellence in Brain Science and Intelligence Technology,Chinese Academy of Sciences, Shanghai, China}
\fntext[c]{School of Future Technology, University of Chinese Academy of Sciences, Beijing, China}
\fntext[d]{National Laboratory of Pattern Recognition, Institute of Automation,Chinese Academy of Sciences, Beijing, China}
\fntext[e]{These authors contributed equally to this work.}
\fntext[f]{Corresponding Author.}

\end{frontmatter}

\section*{SUMMARY}

How neural networks in the human brain represent commonsense knowledge, and complete related reasoning tasks is an important research topic in neuroscience, cognitive science, psychology, and artificial intelligence. 
Although the traditional artificial neural network using fixed-length vectors to represent symbols has gained good performance in some specific tasks, it is still a black box that lacks interpretability, far from how humans perceive the world. Inspired by the grandmother-cell hypothesis in neuroscience, this work investigates how population encoding and spiking timing-dependent plasticity (STDP) mechanisms can be integrated into the learning of spiking neural networks, and how a population of neurons can represent a symbol via guiding the completion of sequential firing between different neuron populations.  The neuron populations of different communities together constitute the entire commonsense knowledge graph, forming a giant graph spiking neural network. Moreover, we introduced the Reward-modulated spiking timing-dependent plasticity (R-STDP) mechanism to simulate the biological reinforcement learning process and completed the related reasoning tasks accordingly, achieving comparable accuracy and faster convergence speed than the graph convolutional artificial neural networks. From a neuroscience perspective, the work in this paper provided the foundation of computational modeling for further exploration of the way the human brain represents commonsense knowledge. For the field of artificial intelligence, this paper indicated the exploration direction for realizing a more robust and interpretable neural network by constructing a commonsense knowledge representation and reasoning spiking neural networks with solid biological plausibility.

\section*{INTRODUCTION}

Commonsense Knowledge representation and reasoning are considered as a key to achieving human-level AI by artificial intelligence researchers~\citep{minsky2007emotion,bisk2020piqa}. After the importance of commonsense knowledge for understanding the physical world was pointed out by Marvin Minsky and John McCarthy, many projects are launched to solve the problem of the absence of commonsense knowledge in the agent of Artificial Intelligence, including Cyc~\citep{lenat1985cyc}, ConceptNet~\citep{liu2004conceptnet}, Machine Common Sense (MCS) ~\citep{shu2021agent}.

With the rise of deep learning methods in recent years, a batch of research on common sense knowledge representation and reasoning based on artificial neural networks(ANN) has emerged. Kocijan et.~\citep{kocijan2019surprisingly} improved the performance of for Winograd schema challenge, which is a popular benchmark for natural commonsense reasoning, by fine-tuning the pre-trained Bidirectional Encoder Representations from Transformers (BERT) Language Model. Yasunaga et. ~\citep{yasunaga2021qa} introduced knowledge graphs utilizing deep graph neural networks for Commonsense Question Answering.
However, data-driven ANN models can only show intelligence at the behavioral level by finding correlations between data. Because these models' information processing methods are far from human cognitive processes, their interpretability is impoverished and can often only be regarded as black boxes.

Spiking Neural networks(SNN), considered as the third generation of the neural networks models~\citep{maass1997networks}, inspired by the structures and functions of the biological neurons~\citep{maass2001pulsed}, have better biological plausibility. SNN has been used in various field including pattern classification ~\citep{zhao2020glsnn}, causal reasoning~\citep{fang2021brain}, sequence information processing~\citep{fang2021SPSNN} and Decision-making~\citep{zhao2018brain,zhao2020neural}.

\begin{figure}
\centering

\includegraphics[width=12cm]{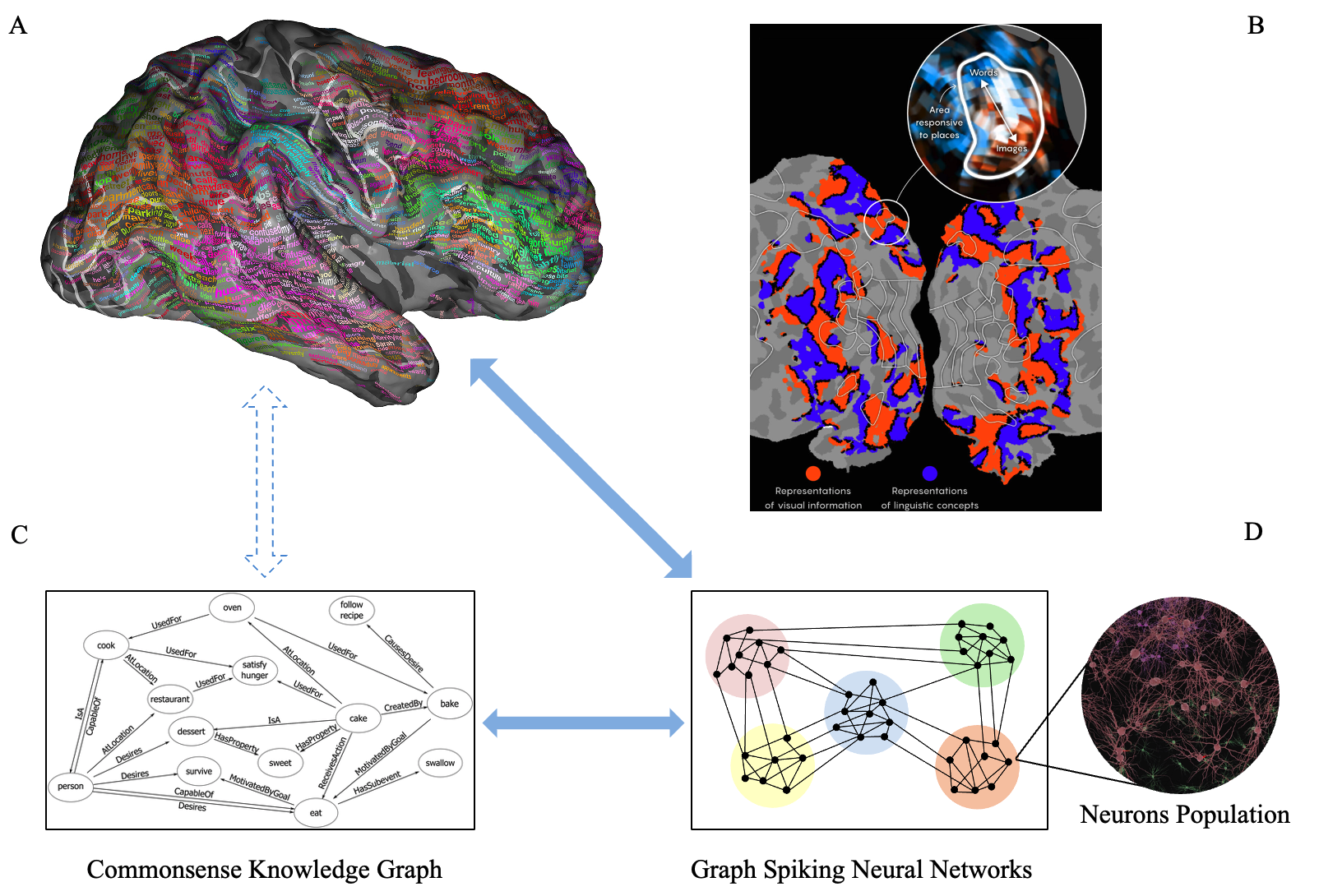}
\caption{ Brain-inspired Graph Spiking Neural Networks represents commonsense knowledge.\\
(A) Semantic maps in the brain, cite from ~\citep{huth2016natural}. \\
(B) Gradients of abstraction in cortex, the visual and linguistic representations of semantic categories align, cite from~\citep{huth2021visual}.\\
(C) Commonsense knowledge graph.\\
(D) Brain-inspired Graph Spiking Neural Networks. Each colored circle represents a neuron population, the small black circles represent neurons, and the line segments represent synaptic connections.   }

\label{whole}
\end{figure}

According to the human cerebral cortex in different positions, there are different degrees of response to different words. Huth et al. have drawn a brain semantic map \cite{huth2016natural} through fMRI research, as figure \ref{whole}A, forming the knowledge graphs in the brain . According to the latest research by Huth et al. in 2021\cite{huth2021visual}, there is a gradual and continuous transition between the visual representation of concepts and the abstract representation at the linguistic level in the human cerebral cortex, forming an abstract gradient in the cortex, as  figure \ref{whole}B. The representation of concepts in the brain is higher than the sensory modality and exists independently, and the semantic relationship network between concepts contains commonsense knowledge\ref{whole}C. Meanwhile, in the deep learning field, Graph Neural Networks（GNN）have been widely used in the field of knowledge representation and reasoning for GNN can encode graph structures, and node attributes compactly ~\citep{schlichtkrull2018modeling,2019Cognitive,zhang2020efficient}. Whereas, few works combine GNN with knowledge representation in the brain.

The grandmother-cell theory is an essential hypothesis about symbolic representation in neuroscience. Its core argument is that a population of neurons in the brain encodes a symbol or concept ~\citep{feldman2009experience,Rodrigo2012Concept,quiroga2019neural}. Recently, neuroscientists discovered that the representation of spatial symbols in the brains of macaque monkeys exists in the form of population coding, further revealing the way symbols are represented in the brains of primates ~\citep{fang2021SPSNN,xie2022geometry}. Inspired by these works, we combine the population coding and STDP mechanism in the spiking neural networks so that the networks can complete the encoding and memory of the commonsense knowledge graph. As figure \ref{whole}D shown, 

Therefore, we proposed graph spiking neural networks for commonsense knowledge representation and reasoning(KRR-GSNN) via fusing Spiking neural networks and Graph neural networks. Besides, we introduced Population coding, Spike Timing Dependent Plasticity (STDP), and Reward-modulated Spike Timing Dependent Plasticity(R-STDP) bio-based brain working mechanisms as the building block of this model, enhancing the biological solidity of our work. Through experimental research, we demonstrated that the spiking neural networks model that integrates brain-inspired mechanisms could complete knowledge representation and related reasoning tasks satisfactorily. Compared with the deep artificial networks model, KRR-GSNN not only has better explainability and biological plausibility but also can converge faster in some reasoning tasks, which excited us intensely.

\section*{RESULTS}

\subsection*{\textbf{ Population Coding for Entities and Relations}}

The triples $ (A \rightarrow R \rightarrow B)  $containing entities and relationships are the basic building block of the knowledge graph. How SNN represent entities, relationships, and connections between them are fundamental and core issues for both artificial intelligence and neuroscience. At the same time, recently, because GNN has better structured information processing capabilities, some researchers have made remarkable progress on a series of natural languages understanding related issues such as question answering system construction~\citep{2020A} and commonsense knowledge reasoning by combining GNN with knowledge graphs~\citep{2021QA}. Then, how to use SNN as the building block of the networks and learn from the working mechanism of the brain and the idea of GNN to achieve better structured information understanding and processing capabilities of the networks has become the core issue we are currently concerned about.

In the human brain, concepts are encoded in the human Medium Temporal Lobe (MTL) by neurons populations called “grandmother-cells,” that respond selectively and invariantly to stimuli representing a specific person or a specific place. ~\citep{quiroga2019neural,quiroga2005invariant,quiroga2012concept} Recently, neuroscience experiments studies have shown that the representation of spatial symbols in the brains of macaque monkeys exists in the form of population coding, further revealing the way symbols are represented in the brains of primates ~\citep{fang2021SPSNN,xie2022geometry}.

Inspired by the working principle of the brain, in this work, we propose to use population coding to represent entities and relationships and use the synaptic connections between populations of neurons to represent the edges in the triad. In the field of computational neuroscience, some scholars have studied how to use the population coding mechanism to realize the representation of different concepts and relationships and learn the synaptic connections between different concepts and relationships through synaptic plasticity~\citep{gastaldi2021shared}. We consider spiking neural networks, which contain N neurons, representing M kinds of concepts and relationships. The pattern  $\varphi ^m = \{\varphi_i^m \in \{0,1\};1 \leq i \leq N\} $ with index $m \in \{1,...,M \}$ represents one of the stored memory engram, i.e. one concept or relationship. A value $\varphi_i^m = 1 $ indicates that neuron $i$ is part of the stored memory engram and therefore belongs to the population of concept $m$, while a value of $\varphi_i^m = 0 $ indicates that it does not. A network that has stored $M$ memory engrams is said to have a memory load of $\alpha = \frac{M}{N}. $ We focus on the sparse coding of memory engrams, i,e, every single neuron i has a low probability $\lambda = Prob(\varphi_i^m =1) \ll 1  $  to participate in the population of concept cells corresponding to memory engram $m$.

In the model of this paper, we introduce the similarity evaluation function $Sim(t)$, i,e, Equation \ref{smi}, to evaluate the representation of different concepts in the networks at a specific moment. 

\begin{equation}\label{smi} 
   Sim^m(t) = \frac{1}{N\lambda(1-\lambda) }\sum_{i=1}^N(\varphi^m_i -\lambda)\sigma _i(t)
\end{equation}

\begin{equation}\label{delta1} 
    \sigma  _i(t)=\left\{
    \begin{array}{rcl}
    0       &      & {V_i(t)<V_{threshold }}\\
    1       &      & {V_i(t)\geq V_{threshold } }
    \end{array} \right.
\end{equation}

In the Equation \ref{smi} and Equation \ref{delta1}, $\sigma (t)$  and $\varphi ^m $ represents networks the firing state and memory engram stored patterns, respectively. $\lambda= Prob(\varphi_i^m =1)  $, representing that single neuron $i$ has a low probability $\lambda$ to participate in the population coding of a memory engram $m$. $N$ is the total number of neurons in the networks.

\begin{figure}
\centering
\setlength{\fboxrule}{0.8pt}
\setlength{\fboxsep}{0.5cm}
\includegraphics[width=12.5cm]{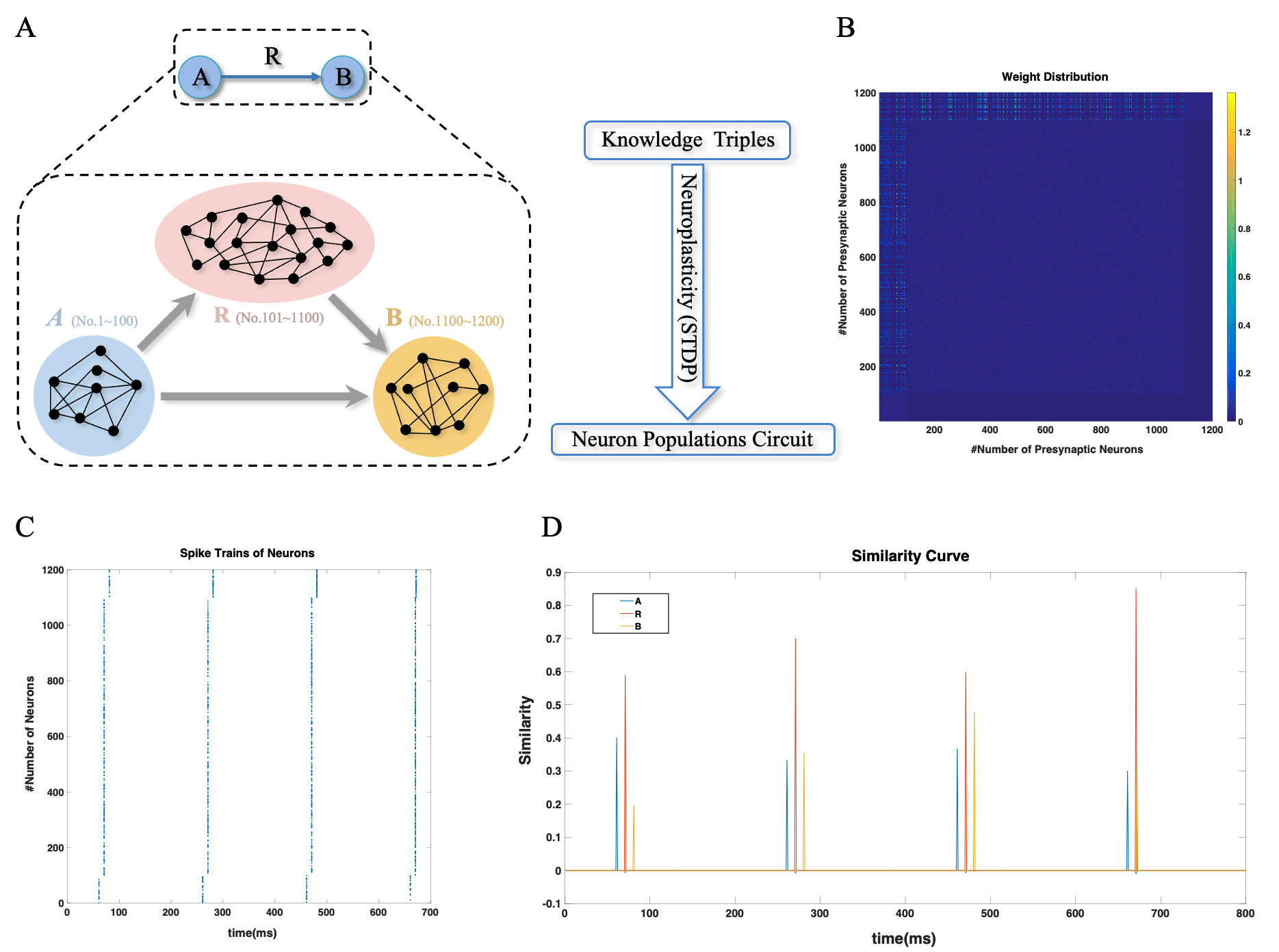}
\caption{Neuron populations representing commonsense knowledge triples.\\
(A)The entities A and B, relation R, correspond to a cluster of neurons, respectively. The gray arrows in the figure represent the synaptic connections between different neuron populations learned through the STDP learning rule, and there are also weak connections within each neuron population simultaneously. \\
(B)The weight distribution of these networks. \\
(C)Figure of neuron spike trains during training and test process.\\
(D)Similarity function curve of the networks during the training and testing process.
}
\label{ARB}
\end{figure}

The similarity measures the correlation between the networks firing state $\sigma (t) $ and the memory engram stored patterns $\varphi ^m$. When the memory engram $m_1$ is retrieved, the neuron population of $m_1$ will firing strongly, shown in Fig.\ref{ARB}A, then  $Sim^{m_1} \approx 1$, and if the whole networks is at resting state, then $Sim^{m} \approx 0$ for every memory engram $m$.

As for the synapse learning rule, Spike Timing Dependent Plasticity (STDP)   ~\citep{bi1998synaptic,dan2004spike} is one of the most crucial learning principles for the biological brain. STDP postulates that the strength of the synapse is dependent on the spike timing difference of the presynaptic and postsynaptic neuron ~\citep{dan2006spike}. Here we use STDP to update synaptic weights according to the relative time between spikes of presynaptic and postsynaptic neurons. The modulation principle is that if the postsynaptic neuron fires a few milliseconds after the presynaptic neuron, the connection between the neurons will be strengthened; otherwise, the connection will be weakened  ~\citep{wittenberg2006malleability}. Please refer to the method chapter for the specific spiking neuron model, synapse update rule, and corresponding parameters.

Inspired by biological discoveries, external input stimulation to the corresponding spike neuron population represents the entities and relations in the knowledge graph. We choose Poisson Encoding as the method of input stimulation. Due to the randomness of the Poisson Encoding,  part of neurons in the population will fire at different times when the external stimulation window is given.

Each neuron population contains a certain proportion of excitatory and inhibitory neurons. In subsequent chapters, we will specifically discuss the influence of the proportion of inhibitory neurons on the experimental effect and the biological interpretability behind it. As shown in Fig.\ref{ARB}, we need to characterize the triple ARB as the connection between different neuron populations in the spiking neural networks and between them. Specifically, A and B are two entities, R represents a relationship, and they are each represented by a population of neurons. When triples like ARB appear in the knowledge graph, we need to give the three neuron populations A, R, and B, in turn, to stimulate the external current and control this process within the STDP window. Due to the plasticity law of STDP, the synaptic connections between the three different neuron populations of the ARB will form, as shown in Fig.\ref{ARB}. The spike trains of networks the learning and testing process are shown in Fig.\ref{ARB}C, and Fig.\ref{ARB}B shows the weight distribution of networks after training. It can be seen in the figure that there are synaptic connections between specific neuron populations, and weak synaptic connections also exist within neuron populations. The change curve of the similarity function in the above process is shown in Fig.\ref{ARB}D. The first three firing processes are the training phase (0-100ms, 200-300ms, 400-500ms), all of which are when external current stimulation is sequentially input to the three neuron populations of the ARB. The networks changes to the Smi of the three ARB memory engrams. 600ms-700ms is the test phase. Only the external currents of the A and R two neuron populations are stimulated. It can be found in Fig.\ref{ARB} D that the similarity of the B neuron population also rises rapidly, thus realizing the memory of the triple ARB by the spiking networks.

After completing the memory of the basic building block, this paper uses this foundation to represent all English triples in ConceptNet as spiking synaptic connections between different neuron populations and complete the follow-up on this basis experiment. For details, please refer to the subsequent chapters.

The above process is to integrate the foundation of the knowledge graph into the learning process of the spiking networks. After completing the learning, the spiking networks can complete the representation and recall of the knowledge graph. The internal connections of different neuron populations and the connections between neuron populations form a semantic networks graph. We call the spiking neural networks that represent the structural knowledge graph as graph spiking neural networks. Based on the encoding method described in this chapter, we encode the English triples in ConceptNet into spiking neural networks, with a total of about 2'500'000 triples, including 17 relationships and about 800'000 entities. The specific experimental results are shown in the next chapter.


\subsection*{\textbf{ Transitive and Non-transitive Relation Reasoning }}

\begin{figure}
\centering
\setlength{\fboxrule}{0.8pt}
\includegraphics[width=12.5cm]{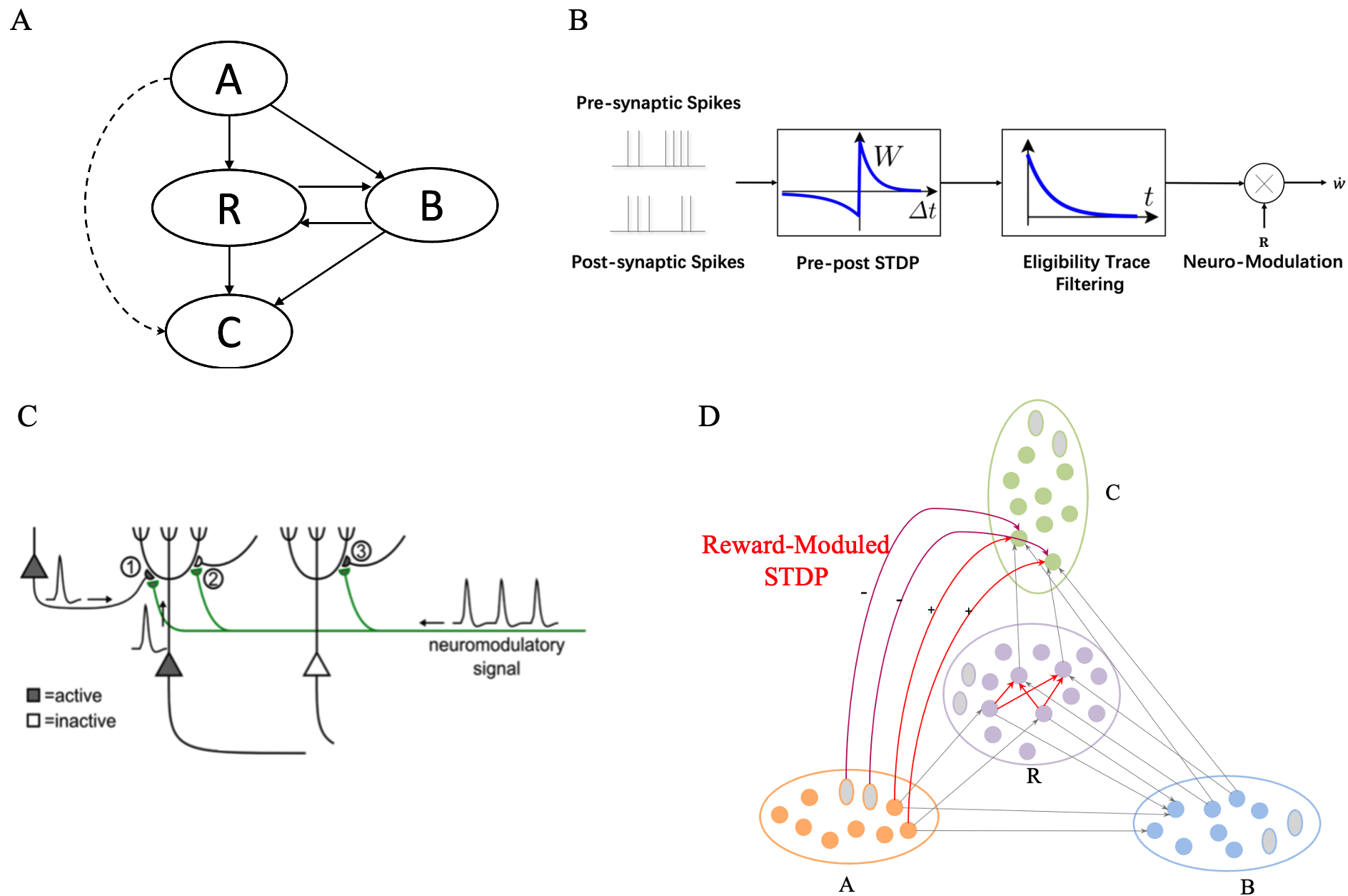}
\caption{Transitive and Non-transitive Relation Learning with Reward-modulated STDP mechanism.\\
(A)Diagram of transitive and non-transitive relation.  \\
(B)Reward-modulated STDP mechanism in computational modeling. \\
(C)Reward-modulated STDP mechanism in the biological brain, cite from~\citep{fremaux2016neuromodulated}.\\
(D)Learning transitive and non-transitive relationships with the R-STDP learning rule. The round neurons represents  excitatory neurons and the oval neurons represents inhibitory neurons.\\ }
\label{RSTDP}
\end{figure}

Different relationships of common sense knowledge graphs have different properties, and one of the essential properties is transitivity, as shown in  Fig.\ref{RSTDP}A. How humans distinguish between transitive and non-transitive relationships at the cognitive level and their differences at the level of neural circuits are important research topics in cognitive science, neuroscience, and psychology.

Based on the method mentioned above to represent triples, we will explore the difference between transitive and non-transitive relationships at the neural circuit level in the graph spiking neural networks fused with common sense knowledge. In the early stage of human cognitive function development, the cognition of the world is continuously improved, mainly in the form of trial and error. Reinforcement learning plays a critical role in this process, including the transitivity of distinguishing relationships~\citep{rovee1969conjugate}.

Therefore, we introduced the Reward-Modulated STDP (R-STDP) (document, mine) mechanism inspired by biology as the learning rule of the networks, as shown in Fig.\ref{RSTDP}C. 
We choose Reward-modulated STDP (R-STDP) to implement the reinforcement learning due to its excellent biology plausibility ~\citep{fremaux2016neuromodulated}. In the process of biological brain learning, the basic STDP plasticity mechanism and reward and punishment signals are integrated through the release of dopamine and other neuromodulators to achieve reinforcement learning~\citep{fremaux2016neuromodulated}. Reward-modulated STDP (R-STDP) mechanism in the computational modeling is shown in Fig.\ref{RSTDP}B. The main idea of R-STDP is to modulate the outcome of original STDP by a reward term~\citep{friedrich2011spatio}. Please see the Methods section for details.

This mechanism uses the eligibility trace function, combines the STDP law and the neuromodulation mechanism as a reward and punishment signal, and can be regarded as an "STDP mechanism with reward and punishment signal." Refer to the chapter on methods for specific mathematical operations.

Going back to the question in this chapter, we used the R-STDP mechanism as a training method and combined it with the Population Coding mechanism to complete the judgment of the transitivity of different relationships. Specifically, we hope the networks can learn whether the \textbf{\textsl{R}} relationship is transitive. After the network learns the \textbf{\textsl{ARB}} and \textbf{\textsl{BRC}} triples, it must determine whether the \textbf{\textsl{ARC}} is established. If the relationship \textbf{\textsl{R}} is a transitive relationship, such as Bigger, then \textbf{\textsl{ARC}} must be established. On the contrary, if R is a non-transitive relationship, such as Anonym, then \textbf{\textsl{ARC}} is not established.

At this time, we will ask the networks whether the \textbf{\textsl{ARC}} is established, and if the networks answer correctly, we will give rewards. Otherwise, we will give punishments. Specifically, the inquiry process of the networks is as follows: We will stimulate the neuron population that characterizes \textbf{\textsl{A}} and \textbf{\textsl{R}}. If \textbf{\textsl{R}} is a  transitive relationship when the Smic value of the networks exceeds the threshold, it means that the networks "answers" \textbf{\textsl{ARC}} and the answer is correct. , We will release the reward signal; on the contrary, if \textbf{\textsl{R}} is a non-transitive relationship when the networks' Smic value exceeds the threshold, and the answer is wrong, we will release the penalty signal. Under the R-STDP learning rule, the synaptic connections within and between neuronal populations will change so that the networks can judge whether ARC is established. As shown in Fig.\ref{RSTDP}D, the neural circuit represented by the red line is the synapse that should be formed when the relationship \textbf{\textsl{R}} is a transitive relationship. The internal connection of the neuron population of the transitive relationship is closer than the non-transitive relationship, and the close connection of the neural circuit supports the agent to complete the transitive reasoning.
Conversely, if the relationship is non-transitive, the internal connections of the neuron population that characterize the relationship will be weaker, and the neural circuits formed by different triads like \textbf{\textsl{ARB}} and \textbf{\textsl{BRC}} will be relatively independent. Moreover, there are excitatory and inhibitory neurons in each neuron population. These two neurons have a competitive relationship in the calculation process. The specific plasticity rules of excitatory neurons and inhibitory neurons are in the Method section. The existence of inhibitory neurons makes the activation of the \textbf{\textsl{A}} neuron population also inhibits the activation of the \textbf{\textsl{C}} neuron population, thereby realizing the agent's negation of the \textbf{\textsl{ARC}} relationship.

\begin{figure}
\centering
\setlength{\fboxrule}{0.8pt}
\fbox{\includegraphics[width=12cm]{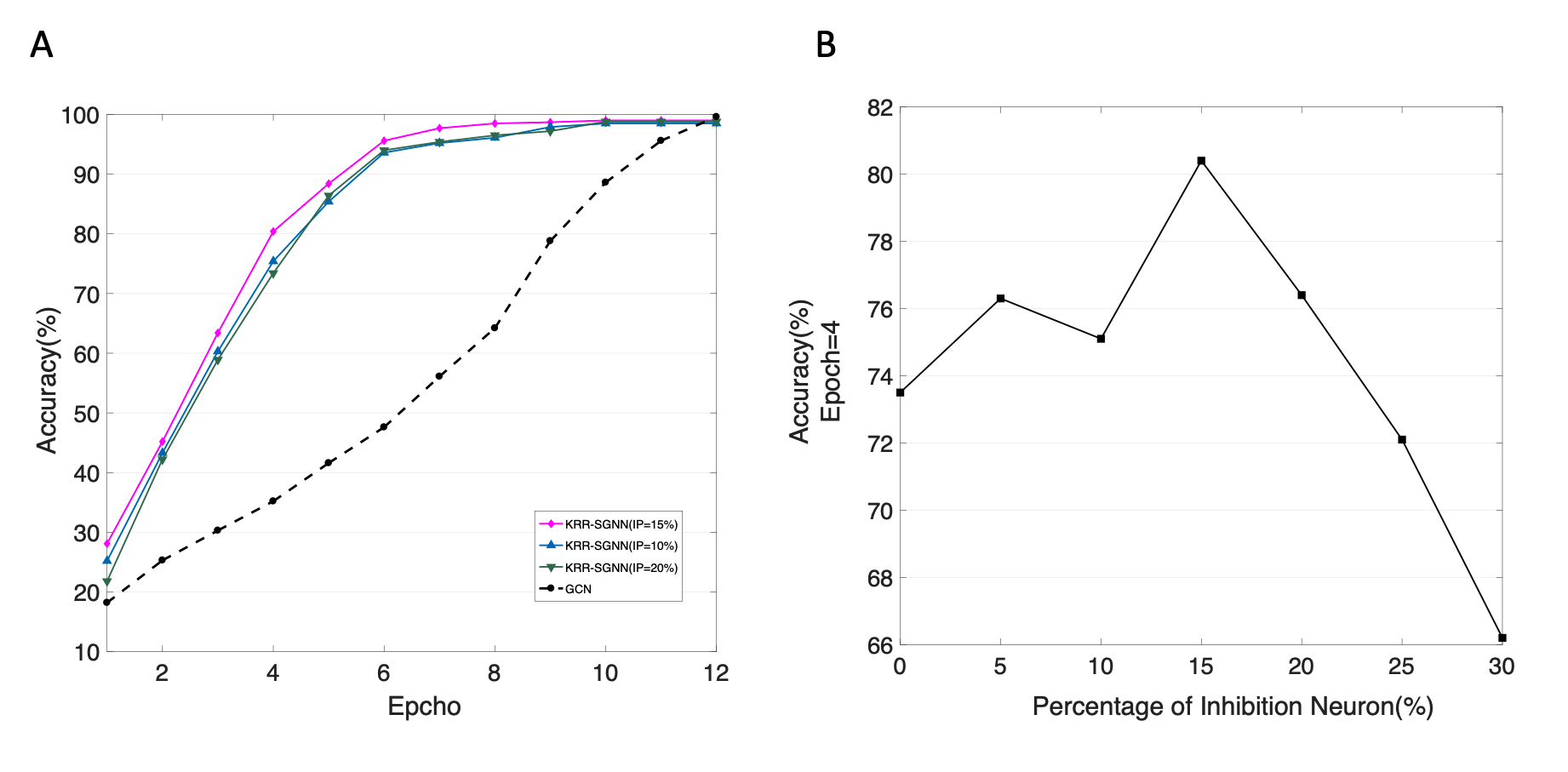}}
\caption{ Relation reasoning results. \\
(A) The figure of the experimental accuracy rate. Our method has gained comparable accuracy and faster convergence speed compared to GCN.\\ 
(B) The figure of the experimental accuracy changes with the proportion of inhibitory neurons. It can be seen that when the proportion of inhibitory neurons is about 15\%, the performance of networks is the best.}

\label{RSTDPres}
\end{figure}

In the actual experiment, we used the English triples in ConceptNet as the experimental data set, including 17 relations, including ten transitive relations, and seven non-transitive relations. Moreover, we clean up all these triples into a graph data set to complete subsequent experiments. Based on 17 different relationships, our entire triplet data set is divided into different sub-graphs. We give 30\% of the triples in the sub-graphs to Mask, thus transforming the entire unsupervised learning process into supervised learning. KRR-GSNN will constantly be "asked" whether specific triples exist and get reward/punishment signals to learn whether a relationship is transitive gradually.
We compared the convergence speed of this model and GCN (Graph Convolution networks) during the training process, and the results are shown in Fig.\ref{RSTDPres}A. We found that although KRR-GSNN is slightly inferior to GCN in terms of final accuracy, the learning speed of KRR-GSNN is much faster than GCN, thus verifying the effectiveness and efficiency of the model.
It is worth mentioning that we compared the effects of different inhibitory neuron ratios in the neuron population on the experimental results. As shown in Fig.\ref{RSTDPres}B, we take the results of different ratios after four iterations as an example. According to the results,  when the proportion of inhibitory neurons is about 15\%, the model's accuracy is the highest. We were surprised to find that this ratio is the same as the CA1 area of ​​the hippocampus, which is responsible for concept learning and new memory form in the human brain ~\citep{nieh2021geometry}.
The proportion of inhibitory neurons is almost the same! Furthermore, the "grandmother cell" is found in the hippocampus area of ​​MTL, which we will discuss in detail in the subsequent discussion chapters.

\subsection*{\textbf{ Generating Conceptual Commonsense Knowledge through Entity Conceptualization} }

\begin{figure}
\centering
\includegraphics[width=12.5cm]{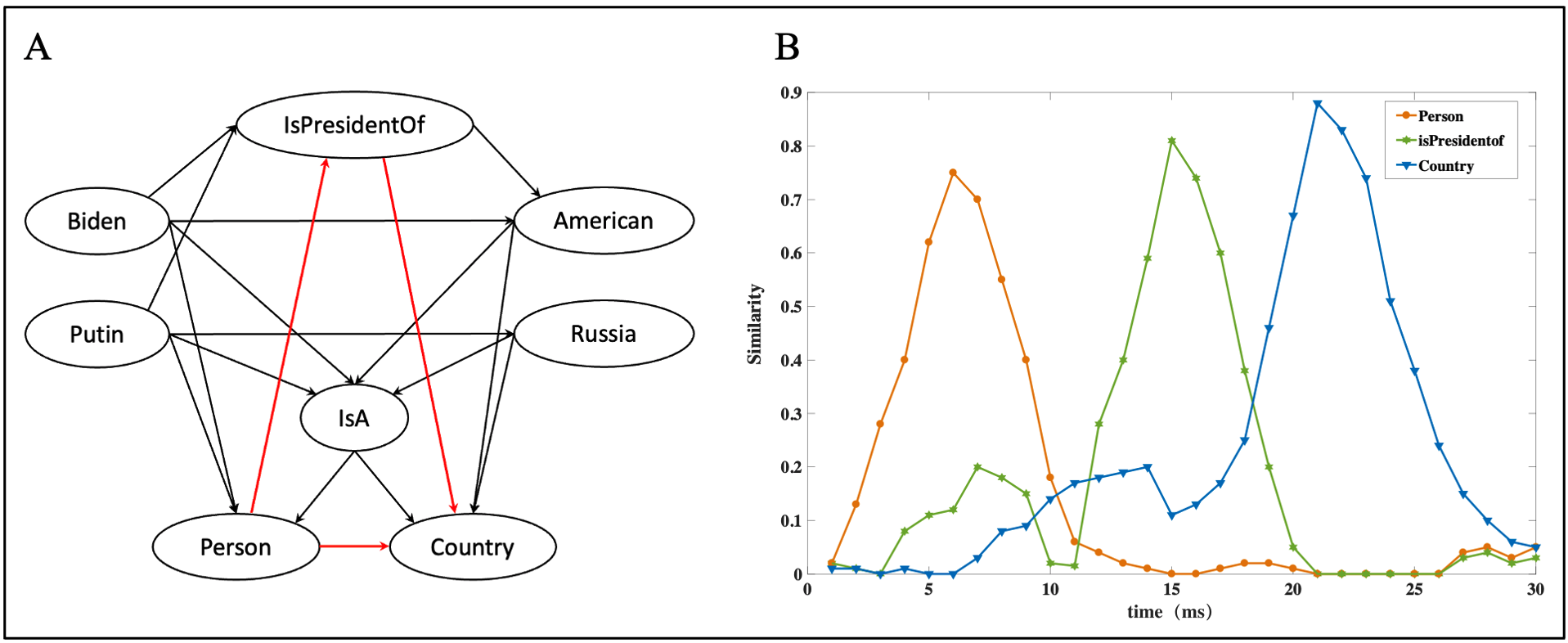}
\caption{Conceptual Commonsense Knowledge Generating\\
(A)Instance of conceptual commonsense knowledge generating.\\
(B)Similarity function curve of the networks during knowledge generating.
}
\label{KG}
\end{figure}
    
Many sentences describe the relations between two objects in the commonsense knowledge bases, which could be expanded by extracting more abstract knowledge.
This section will demonstrate spiking networks' ability to generate higher-level abstract knowledge by encoding common sense knowledge into GSNN. We conducted a specific analysis on an example, which is shown in Figure\ref{KG}A. Each ellipse in Figure\ref{KG}A represents an entity or relationship corresponding to a population of spiking neurons with a specific connection structure. The synaptic connections (black connections) between them are learned through the STDP law, and the process is as Result1. The method described in the chapter accumulates multiple triples. The specific triples contained in the figure are as follows:\\
\textit{
(a). Biden is the president of America.\\
(b). Putin is the president of Russia.\\
(c). Biden is a person.\\
(d). Putin is a person.\\
(e). America is a country.\\
(f). Russia is a country.\\
} 
Through the above knowledge, a more abstract knowledge can be generated by reasoning:\\
\textit{A person is the president of a country.}\\
The corresponding synaptic connection is shown in the red directional connection in Figure\ref{KG}A.
 
For this example, we try to let the networks learn a-f knowledge, they corresponding to the six triples in the experiment, and then GSNN acquires the more abstract conceptual knowledge of \textsl{A person is the president of a country } through autonomous induction. 

In the specific experiment process, we firstly coded six pieces of knowledge into GSNN through the STDP learning rules through the method in the Result1 chapter and obtained all the black synaptic connections in Figure\ref{KG}A. Then continue to stimulate the \textsl{Biden} and \textsl{Putin} neuron populations simultaneously so that the networks can begin to summarize the "common ground". Due to the continuous activation of the \textsl{Biden} and \textsl{Putin} neuron populations, the \textsl{Person}, \textsl{IsA}, and \textsl{IsPresidentOf} neuron populations are also gradually activated. It is worth mentioning that because there will also be a connection between \textsl{IsA} and \textsl{Person}, the \textsl{Person} neuron population will be activated first. 

Furthermore, because \textsl{Biden} and \textsl{IsPresidentOf} are activated, the \textsl{America} neuron population is activated, and the \textsl{Russia} neuron population is activated too. The activation of \textsl{America}, \textsl{Russia}, and \textsl{IsA} neuron populations directly leads to the activation of \textsl{Country}. At this point, we found that the \textsl{Person}, \textsl{IsPresidentOf}, and \textsl{Country} neuron populations are activated sequentially. According to the STDP rule, new synaptic connections will also be formed that fire within the time window. To our surprise, we only stimulated two neuron populations, \textsl{Biden} and \textsl{Putin}, and KRR-GSNN altogether concluded the knowledge that  \textsl{A person is the president of a country } without other intervention. The Similarity function change process of this process networks is shown in Figure\ref{KG}B.\\
    
After the new synaptic connection is established, we try to verify whether GSNN has learned more abstract new knowledge, so we stimulate the \textsl{Person} and \textsl{IsPresidentOf} neuron populations and then find that the \textsl{Country} neuron population is activated. This result shows that the networks has indeed acquired this new knowledge through self-inductive learning.\\
 
It should be emphasized that through this experiment, we are not trying to prove that KRR-GSNN has the ability to generate universally correct new knowledge for the entire data set but to explore the internal working mechanism of humans, especially infants, in the process of inductive learning. One possible mode of operation. For this example, after encoding the existing knowledge graph into GSNN, we only stimulate the two concepts of \textsl{Biden} and \textsl{Putin}. The networks can autonomously generate inductive learning behavior and successfully generate correct knowledge. At present, in neuroscience, the process of human inductive generation of new knowledge is still in an unknown state. Our research provides the possibility of a computational model for this process.

\section*{Discussion}
In the current work, we have completed the representation learning of common sense knowledge and related reasoning tasks based on the pulse graph neural networks. In this process, we used various biologically based brain working mechanisms such as Population Coding, STDP, and R-STDP, which proved the feasibility of using spiking networks to complete commonsense knowledge representation and reasoning. As far as we know, this work is the first work that uses spiking networks for common sense knowledge representation and reasoning, and it is also the first work related to graph spiking neural networks.

Spiking neural networks in commonsense knowledge representation and reasoning has a reference significance in constructing a brain-inspired system of semantic understanding. Computational modeling will help us to understand how to use the essential elements of biological neurons, rules of synaptic learning, and spiking neural networks to deal with common sense knowledge~\citep{davis2015commonsense}. It provides a computational foundation for exploring how the human brain encodes commonsense knowledg e and performing reasoning.

Using neural networks to implement symbolic reasoning has always been an urgent problem in artificial intelligence. The human brain perfectly realizes the use of neural networks for symbolic reasoning. The problem of poor interpretability of artificial neural networks under connectionism in intelligence is essential because the way it processes information is far from the "thinking" process of the human brain. We hope to learn from the brain, based on the structure and working mechanism of related neural circuits in the brain, to construct a structurally and functionally brain-inspired and behaviorally human-like spiking neural networks model for symbolic reasoning.

In recent years, evidence of the "grandmother cell" theory in the field of experimental neuroscience has been continuously discovered ~\citep{Rodrigo2012Concept,xie2022geometry}, and only a few works ~\citep{fang2021SPSNN} to apply this theory to functional intelligent systems. Inspired by the theory of "grandmother cells" in neuroscience, we propose using a population of neurons to represent a specific concept or relationship, which we call Memory Engram. In order to better observe the discharge of the networks, we propose the similarity function to show how the networks represent different concepts or relationships at different times. Through the similarity function, we can trace the reasoning path of the spiking networks in the reasoning process, which significantly improves the interpretability of the networks. KRR-GSNN is no longer a black box, and we can observe how the networks complete the inference process, which also avoids the problem of behaviorism. Whether the result is correct is no longer the only indicator to measure the quality of the model. We can also control the reasoning process of getting the result, which can enhance the robustness of the model. However, from the perspective of artificial intelligence ethics and safety, it is also of great significance to construct artificial intelligence that is more understandable and controllable by humans.

One step further, we represented many commonsense knowledge graphs into KRR-GSNN and constructed specific reasoning tasks for the transitivity of relationships. Utilizing self-supervised reinforcement learning, we trained KRR-GSNN to implement the basic reasoning process and obtained a faster convergence rate than GCN (Graph Convolutional networks). Moreover, by adjusting the proportion of inhibitory neurons in the neuron population that characterizes a Memory Engram, we found that when the proportion of inhibitory neurons is about 15\%, the accuracy rate under the same number of iterations is the highest. What surprises us is that the CA1 area of ​​the hippocampus that is responsible for concept learning and new memory formation in the human brain, which is the brain area where "grandmother cells" are found, is precisely 14-15\% of inhibitory neurons ~\citep{nieh2021geometry}. Perhaps our process of adjusting the proportion of inhibitory neurons in KRR-GSNN is also in line with the process of constant trial and error in the evolution of the brain. In the end, the ratios are very close, which also shows that our model has solid biological rationality.

After KRR-GSNN completed the coding of many common-sense knowledge graphs, we continued to explore the model's inductive ability for conceptualizing common sense knowledge and completed some specific examples of inductive reasoning. Through this experiment, we hope to explore possible working mechanisms and patterns in the brain of humans, especially infants, in the process of inductive learning. For a specific instance, after encoding the existing knowledge graph into GSNN, KRR-GSNN can autonomously generate inductive learning behaviors and successfully induce new correct knowledge. At present, in neuroscience, the process of human inductive generation of new knowledge is still unknown. This research provides the possibility of a computational model for follow-up neuroscience and cognitive science-related research.

In summary, we are based on spiking neural networks. We use various biologically based brain working mechanisms such as Population Coding, STDP, R-STDP., which proves the feasibility of using spiking networks to complete common sense knowledge representation and reasoning. In addition, by introducing the similarity function to monitor the network's status, the reasoning path of the networks can be obtained, making KRR-GSNN far more interpretable than traditional ANN, more in line with AI ethics, more secure, and controllable. In the deductive reasoning related to the transitivity of the relationship, our model exhibits a faster convergence rate and extremely close final accuracy than GCN. 
 

\section*{Limitations of the study}
KRR-GSNN is a pioneering exploration of commonsense knowledge representation and reasoning based on spiking neural network, and there must be some deficiencies. \\
Firstly, KRR-GSNN emphasizes the representation of knowledge at the symbolic level and lacks insight into the role of multimodal sensory input in commonsense knowledge. Although KRR-GSNN can complete symbolic reasoning based on spiking neural network, it integrates the ideas of symbolic and connectionist artificial intelligence to a certain extent \cite{hu2016harnessing}, but due to the lack of insight into embodied intelligence \cite{shapiro2010embodied,wilson2002six}, this directly leads to the complexity of the reasoning KRR-GSNN  can complete is still at a low level, and there is still a big gap with the reasoning capability of humans.\\shapiro2010embodied,wilson2002six

Furthermore, KRR-GSNN does not take into account the spatial relationship between groups of neurons representing different concepts in the model, but stores them in the form of indexes. The human cerebral cortex has clear regional divisions for the representation of different concepts and knowledge, forming the semantic map\cite{huth2016natural}. In future work, how to combine the knowledge representation completed by KRR-GSNN with the semantic map  in the brain discovered by neuroscientists will be the key issue.

\section*{FUNDING}
This work was supported by the new generation of artificial intelligence major project of the Ministry of Science and Technology of the People's Republic of China (Grant No. 2020AAA0104305), the Strategic Priority Research Program of the Chinese Academy of Sciences (Grant No. XDB32070100),the Beijing Municipal Commission of Science and Technology (Grant No. Z181100001518006).
 

\section*{ACKNOWLEDGMENTS}
For valuable discussions, the authors appreciate Feifei Zhao, Jinyu Fan, Dongcheng Zhao, Qian Zhang, Cunqing Huangfu, and Qian Liang. The authors would like to thank all the reviewers for their help in shaping and refining the paper.

\section*{REFERENCES}
   
\bibliography{KRRGSNN}

\begin{thebibliography}{10}
\expandafter\ifx\csname url\endcsname\relax
  \def\url#1{\texttt{#1}}\fi
\expandafter\ifx\csname urlprefix\endcsname\relax\def\urlprefix{URL }\fi
\expandafter\ifx\csname href\endcsname\relax
  \def\href#1#2{#2} \def\path#1{#1}\fi

\bibitem{minsky2007emotion}
M.~Minsky, The emotion machine: Commonsense thinking, artificial intelligence,
  and the future of the human mind, Simon and Schuster, 2007.

\bibitem{bisk2020piqa}
Y.~Bisk, R.~Zellers, J.~Gao, Y.~Choi, et~al., Piqa: Reasoning about physical
  commonsense in natural language, in: Proceedings of the AAAI Conference on
  Artificial Intelligence, Vol.~34, 2020, pp. 7432--7439.

\bibitem{lenat1985cyc}
D.~B. Lenat, M.~Prakash, M.~Shepherd, Cyc: Using common sense knowledge to
  overcome brittleness and knowledge acquisition bottlenecks, AI magazine 6~(4)
  (1985) 65--65.

\bibitem{liu2004conceptnet}
H.~Liu, P.~Singh, Conceptnet—a practical commonsense reasoning tool-kit, BT
  technology journal 22~(4) (2004) 211--226.

\bibitem{shu2021agent}
T.~Shu, A.~Bhandwaldar, C.~Gan, K.~A. Smith, S.~Liu, D.~Gutfreund, E.~Spelke,
  J.~B. Tenenbaum, T.~D. Ullman, Agent: A benchmark for core psychological
  reasoning, arXiv preprint arXiv:2102.12321 (2021).

\bibitem{kocijan2019surprisingly}
V.~Kocijan, A.-M. Cretu, O.-M. Camburu, Y.~Yordanov, T.~Lukasiewicz, A
  surprisingly robust trick for winograd schema challenge, arXiv preprint
  arXiv:1905.06290 (2019).

\bibitem{yasunaga2021qa}
M.~Yasunaga, H.~Ren, A.~Bosselut, P.~Liang, J.~Leskovec, Qa-gnn: Reasoning with
  language models and knowledge graphs for question answering, arXiv preprint
  arXiv:2104.06378 (2021).

\bibitem{maass1997networks}
W.~Maass, Networks of spiking neurons: the third generation of neural network
  models, Neural networks 10~(9) (1997) 1659--1671.

\bibitem{maass2001pulsed}
W.~Maass, C.~M. Bishop, Pulsed neural networks, MIT press, 2001.

\bibitem{zhao2020glsnn}
D.~Zhao, Y.~Zeng, T.~Zhang, M.~Shi, F.~Zhao, Glsnn: A multi-layer spiking
  neural network based on global feedback alignment and local stdp plasticity,
  Frontiers in Computational Neuroscience 14 (2020).

\bibitem{fang2021brain}
H.~Fang, Y.~Zeng, A brain-inspired causal reasoning model based on spiking
  neural networks, in: 2021 International Joint Conference on Neural Networks
  (IJCNN), IEEE, 2021, pp. 1--5.

\bibitem{fang2021SPSNN}
H.~Fang, Y.~Zeng, F.~Zhao, Brain inspired sequences production by spiking
  neural networks with reward-modulated stdp, Frontiers in Computational
  Neuroscience 15 (2021) 8.

\bibitem{zhao2018brain}
F.~Zhao, Y.~Zeng, B.~Xu, A brain-inspired decision-making spiking neural
  network and its application in unmanned aerial vehicle, Frontiers in
  neurorobotics 12 (2018) 56.

\bibitem{zhao2020neural}
F.~Zhao, Y.~Zeng, A.~Guo, H.~Su, B.~Xu, A neural algorithm for drosophila
  linear and nonlinear decision-making, Scientific Reports 10~(1) (2020) 1--16.

\bibitem{huth2016natural}
A.~G. Huth, W.~A. De~Heer, T.~L. Griffiths, F.~E. Theunissen, J.~L. Gallant,
  Natural speech reveals the semantic maps that tile human cerebral cortex,
  Nature 532~(7600) (2016) 453--458.

\bibitem{huth2021visual}
S.~F. Popham, A.~G. Huth, N.~Y. Bilenko, F.~Deniz, J.~S. Gao, A.~O.
  Nunez-Elizalde, J.~L. Gallant, Visual and linguistic semantic representations
  are aligned at the border of human visual cortex, Nature neuroscience 24~(11)
  (2021) 1628--1636.

\bibitem{schlichtkrull2018modeling}
M.~Schlichtkrull, T.~N. Kipf, P.~Bloem, R.~Van Den~Berg, I.~Titov, M.~Welling,
  Modeling relational data with graph convolutional networks, in: European
  semantic web conference, Springer, 2018, pp. 593--607.

\bibitem{2019Cognitive}
M.~Ding, C.~Zhou, Q.~Chen, H.~Yang, J.~Tang, Cognitive graph for multi-hop
  reading comprehension at scale, in: ACL, 2019.

\bibitem{zhang2020efficient}
Y.~Zhang, X.~Chen, Y.~Yang, A.~Ramamurthy, B.~Li, Y.~Qi, L.~Song, Efficient
  probabilistic logic reasoning with graph neural networks, arXiv preprint
  arXiv:2001.11850 (2020).

\bibitem{feldman2009experience}
V.~Feldman, L.~G. Valiant, Experience-induced neural circuits that achieve high
  capacity, Neural computation 21~(10) (2009) 2715--2754.

\bibitem{Rodrigo2012Concept}
R.~Q. Quiroga, Concept cells: the building blocks of declarative memory
  functions, Nature Reviews Neuroscience 13~(8) (2012) 587--97.

\bibitem{quiroga2019neural}
R.~Q. Quiroga, Neural representations across species, Science 363~(6434) (2019)
  1388--1389.

\bibitem{xie2022geometry}
Y.~Xie, P.~Hu, J.~Li, J.~Chen, W.~Song, X.-J. Wang, T.~Yang, S.~Dehaene,
  S.~Tang, B.~Min, et~al., Geometry of sequence working memory in macaque
  prefrontal cortex, Science 375~(6581) (2022) 632--639.

\bibitem{2020A}
S.~Ji, S.~Pan, E.~Cambria, P.~Marttinen, P.~S. Yu, A survey on knowledge
  graphs: Representation, acquisition and applications (2020).

\bibitem{2021QA}
M.~Yasunaga, H.~Ren, A.~Bosselut, P.~Liang, J.~Leskovec, Qa-gnn: Reasoning with
  language models and knowledge graphs for question answering, 2021.

\bibitem{quiroga2005invariant}
R.~Q. Quiroga, L.~Reddy, G.~Kreiman, C.~Koch, I.~Fried, Invariant visual
  representation by single neurons in the human brain, Nature 435~(7045) (2005)
  1102--1107.

\bibitem{quiroga2012concept}
R.~Q. Quiroga, Concept cells: the building blocks of declarative memory
  functions, Nature Reviews Neuroscience 13~(8) (2012) 587--597.

\bibitem{gastaldi2021shared}
C.~Gastaldi, T.~Schwalger, E.~De~Falco, R.~Q. Quiroga, W.~Gerstner, When shared
  concept cells support associations: theory of overlapping memory engrams,
  bioRxiv (2021).

\bibitem{bi1998synaptic}
G.-q. Bi, M.-m. Poo, Synaptic modifications in cultured hippocampal neurons:
  dependence on spike timing, synaptic strength, and postsynaptic cell type,
  Journal of neuroscience 18~(24) (1998) 10464--10472.

\bibitem{dan2004spike}
Y.~Dan, M.-m. Poo, Spike timing-dependent plasticity of neural circuits, Neuron
  44~(1) (2004) 23--30.

\bibitem{dan2006spike}
Y.~Dan, M.-M. Poo, Spike timing-dependent plasticity: from synapse to
  perception, Physiological reviews 86~(3) (2006) 1033--1048.

\bibitem{wittenberg2006malleability}
G.~M. Wittenberg, S.~S.-H. Wang, Malleability of spike-timing-dependent
  plasticity at the ca3--ca1 synapse, Journal of Neuroscience 26~(24) (2006)
  6610--6617.

\bibitem{rovee1969conjugate}
C.~K. Rovee, D.~T. Rovee, Conjugate reinforcement of infant exploratory
  behavior, Journal of experimental child psychology 8~(1) (1969) 33--39.

\bibitem{fremaux2016neuromodulated}
N.~Fr{\'e}maux, W.~Gerstner, Neuromodulated spike-timing-dependent plasticity,
  and theory of three-factor learning rules, Frontiers in neural circuits 9
  (2016) 85.

\bibitem{friedrich2011spatio}
J.~Friedrich, R.~Urbanczik, W.~Senn, Spatio-temporal credit assignment in
  neuronal population learning, PLoS Comput Biol 7~(6) (2011) e1002092.

\bibitem{nieh2021geometry}
E.~H. Nieh, M.~Schottdorf, N.~W. Freeman, R.~J. Low, S.~Lewallen, S.~A. Koay,
  L.~Pinto, J.~L. Gauthier, C.~D. Brody, D.~W. Tank, Geometry of abstract
  learned knowledge in the hippocampus, Nature (2021) 1--5.

\bibitem{davis2015commonsense}
E.~Davis, G.~Marcus, Commonsense reasoning and commonsense knowledge in
  artificial intelligence, Communications of the ACM 58~(9) (2015) 92--103.

\bibitem{hu2016harnessing}
Z.~Hu, X.~Ma, Z.~Liu, E.~Hovy, E.~Xing, Harnessing deep neural networks with
  logic rules, arXiv preprint arXiv:1603.06318 (2016).

\bibitem{shapiro2010embodied}
L.~Shapiro, Embodied cognition, Routledge, 2010.

\bibitem{wilson2002six}
M.~Wilson, Six views of embodied cognition, Psychonomic bulletin \& review
  9~(4) (2002) 625--636.

\bibitem{HodgkinA}
A.~L. Hodgkin, A.~F. Huxley, A quantitative description of membrane current and
  its application to conduction and excitation in nerve., Bulletin of
  Mathematical Biology 52~(1-2) (1952) 25--71.

\bibitem{miller2018introductory}
P.~Miller, An Introductory Course in Computational Neuroscience, MIT Press,
  2018.

\bibitem{izhikevich2003simple}
E.~M. Izhikevich, Simple model of spiking neurons, IEEE Transactions on neural
  networks 14~(6) (2003) 1569--1572.

\end{thebibliography}

\newpage

\centerline{\textbf{Supplementary Materials}}

\section*{Methods}
    \subsection*{\textbf{ Neuron Model}}
    The building block of our spiking neural networks is the spike neuron model. There are various neuron models such as the famous HH model~\citep{HodgkinA}, Leaky Integrate-and-Fire neuron ({\it LIF}) model~\citep{miller2018introductory}, Izhikevich neuron model~\citep{izhikevich2003simple}, and so on.

\begin{equation}\label{CM}
    C_m\frac{dV}{dt} = -g(V-V_s)+I 
\end{equation}

\begin{equation}\label{tau}
   \tau_m \frac{dV}{dt} = -(V-V_s)+ \frac{I}{g} 
\end{equation}

\begin{equation}\label{reset}
    V\rightarrow V_{reset}, \quad if (V \geq V_{threshold}).
\end{equation}

In order to balance the computational complexity of the model, we choose the Leaky Integrate-and-Fire ({\it LIF}) neuron model as the building block of the Spiking Neural networks. Standard LIF models are shown in Equation \ref{CM} , Equation \ref{tau}  and Equation \ref{reset}. $C_m$ is the membrane capacitance of the neuron, $V$ is the membrane potential of the neuron, $g$ is the conductance of the membrane, $V_s$ is the steady-state leaky potential, here we let $V_s = V_{reset}$ to simplify the model. $I$ is the input current of the neuron. $\tau_m  = \frac{C_m}{g}$ represents the voltage delay time, and different types of neurons have different values of $\tau_m$.

\begin{equation}\label{eq4} 
   I = \sum \limits_{j} w_{j,i} \sigma _j(t-1) + I_s
\end{equation}

\begin{equation}\label{delta} 
    \sigma  _i(t)=\left\{
    \begin{array}{rcl}
    0       &      & {V<V_{threshold }}\\
    1       &      & {V\geq V_{threshold } }
    \end{array} \right.
\end{equation}

Equation \ref{eq4} shows that the current of neurons consists of two parts: the current from other neurons and the external stimulating current $I_s$. $W_{j,i}$ is the weight of i-th neuron to j-th neuron. $\sigma _i(t)$ is the indicator to judge if the $i$-th neuron firing at the time of $t$ in Equation \ref{delta}. \\

All the parameters can be found in Table \ref{table_1}  . \\

\begin{table}\centering
\begin{tabular}{|llllllllll|}
\hline

Model/Rule          &    &   &           & Parameter          &   &  &    &       & Value               \\ \hline
LIF model           &    &   &           & $C_m  $            &   &  &    &       & $30  nF$         \\
                    &    &   &           & $\tau _m  $        &   &  &    &       & 30ms                \\
                    &    &   &           & $V_{reset}$        &   &  &    &       & -65mv                \\
                    &    &   &           & $V_{threshold} $   &   &  &    &       & -35mv                \\

                    &    &   &           & $\tau _{ref}$      &   &  &    &       & 10ms                \\
STDP Rule           &    &   &           & $\tau _s$          &   &  &    &       & 30ms                \\
                    &    &   &           & $\tau _w$          &   &  &    &       & 20ms                \\
                    &    &   &           & $A _+$             &   &  &    &       & 1.1                  \\
                    &    &   &           & $A _-$             &   &  &    &       & 0.95                \\ 
R-STDP Rule         &    &   &           & $C_r$              &   &  &    &       & 10         \\
                    &    &   &           & $C_p$              &   &  &    &       & -10        \\
                    &    &   &           & $T_R$              &   &  &    &       & 5ms        \\   \hline

\end{tabular}

\caption{\textbf{Model parameters.} }
\label{table_1}
\end{table}

\subsection*{\textbf{ Neuroplasticity Rule: STDP and R-STDP}}

As for the synapse learning rule, Spike Timing Dependent Plasticity (STDP)   ~\citep{bi1998synaptic,dan2004spike} is one of the most crucial learning principles for the biological brain. STDP postulates that the strength of the synapse is dependent on the spike timing difference of the pre- and post-neuron ~\citep{dan2006spike}.

Here we use STDP to update synaptic weights according to the relative time between spikes of presynaptic and postsynaptic neurons. The modulation principle is that if the postsynaptic neuron fires a few milliseconds after the presynaptic neuron, the connection between the neurons will be strengthened; otherwise, the connection will be weakened  ~\citep{wittenberg2006malleability}.
The update function is shown in Equation \ref{eq7}, where $A _+$ and $A _−$ are learning rates. $\tau _s$ and $\tau _w$ are STDP time constant, and $\Delta t $ is the delay time from the presynaptic spike to the postsynaptic spike. \\
\begin{equation}\label{eq7}
    \Delta w_{j,i} =\left\{
    \begin{array}{rcl}
    &A_+e^{( \Delta t/\tau _s )}      \quad & { -\tau_w <\Delta t <0 }\\
    &-A_-e^{(-\Delta t/\tau _s )}     \quad & { 0<\Delta t< \tau_w }
    \end{array} \right.
\end{equation}

Besides STDP, we choose Reward-modulated STDP (R-STDP) to implement the reinforcement learning due to its excellent biology plausibility ~\citep{fremaux2016neuromodulated}. In the process of biological brain learning, the basic STDP plasticity mechanism and reward and punishment signals are integrated through the release of dopamine and other neuromodulators to achieve reinforcement learning~\citep{fremaux2016neuromodulated}. Reward-modulated STDP (R-STDP) is the computational modeling of this process. The main idea of R-STDP is to modulate the outcome of 'standard' STDP by a reward signal~\citep{friedrich2011spatio}. 
Synaptic eligibility trace, shown in figure \ref{RSTDP}B,  stores a temporary memory of the STDP outcome so that it is still available by the time a delayed reward signal is received~\citep{fremaux2016neuromodulated}. We regard the timing condition (or 'learning window') of traditional STDP as $STDP(n_i, n_j) $, $n_i$ and $n_j$ denote the presynaptic and postsynaptic neuron in the networks. The synaptic eligibility trace keeps a transient memory in the form of a running average of recent spike-timing coincidences. Synaptic eligibility traces arise from theoretical considerations and effectively bridge the temporal gap between the neural activity and the reward signal.\\
\begin{equation}\label{eq7}
   \Delta e_{j,i} = - \frac{e_{j,i} }{\tau _e} + STDP(n_i, n_j) 
\end{equation}

$e_{j,i}$ is the eligibility traces between presynaptic neuron i and postsynaptic neuron j, $\tau _e$ is the time constant of the eligibility trace. The running average is equivalent to a low-pass filter.       
In R-STDP mechanism, the synaptic weight $W$ changes when the neuromodulator $M$ signals exist. 
\begin{equation}\label{eq8}
   \Delta W = R * E
\end{equation}

Considering the complexity of the networks, we simply choose R-max policy, $R$ is the reward or punish signal towards networks which is given by the experiment environment. Actually, $R$ is the function of time $t$, Equation \eqref{eq9} shows how $R$ changes through time.
\begin{equation}\label{eq9}
   R(t) =\left\{
        \begin{array}{rcl}
        &C_r      \quad &  t - t_r  \leq T_R  \\
        &C_p      \quad &  t - t_p  \leq T_R\\
        &0        \quad &  otherwise
        \end{array} \right.
\end{equation}
$C_r$ and $C_p$ are the constants of reward and punish signal. $t_r$ and $t_p$ denote the latest time of reward and punish. And $T_R$ is the size of time window of reward or punish signal. In the experiment, we set $C_r=10$, $C_p=-10$, and $T_R=5$.\\

All the parameters can be found in Table \ref{table_1}. \\

\end{document}